\documentclass[10pt, conference, compsocconf]{IEEEtran}
\IEEEoverridecommandlockouts
\usepackage{cite}
\usepackage{amsmath,amssymb,amsfonts}
\usepackage{algorithmic}
\usepackage{graphicx}
\usepackage{textcomp}
\usepackage{multirow}
\usepackage[normalem]{ulem}
\usepackage[table,xcdraw]{xcolor}
\usepackage{graphicx}
\usepackage{subcaption}
\usepackage{hyperref}

\useunder{\uline}{\ul}{}

\def\BibTeX{{\rm B\kern-.05em{\sc i\kern-.025em b}\kern-.08em
    T\kern-.1667em\lower.7ex\hbox{E}\kern-.125emX}}
\begin{document}

\title{Hybrid Feature Learning for Handwriting Verification}
\author{\IEEEauthorblockN{
Seyed Mohammad Abuzar Hashemi\IEEEauthorrefmark{1},
Mihir Chauhan\IEEEauthorrefmark{2}, 
Jun Chu\IEEEauthorrefmark{3} and
Sargur Srihari\IEEEauthorrefmark{4}}
\IEEEauthorblockA{The Department of Computer Science and Engineering\\
The State University of New York at Buffalo\\
Buffalo, NY, USA\\
Email: \IEEEauthorrefmark{1}syed.mah.313@gmail.com,
\IEEEauthorrefmark{2}mihirhem@buffalo.edu,
\IEEEauthorrefmark{3}jchu6@buffalo.edu,
\IEEEauthorrefmark{5}srihari@cedar.buffalo.edu}}
\maketitle

\begin{abstract}
We propose an effective Hybrid Deep Learning (HDL) architecture for the task of determining the probability that a questioned handwritten word has been written by a known writer. HDL is an amalgamation of Auto-Learned Features (ALF) and Human-Engineered Features (HEF). To extract auto-learned features we use two methods: First, Two Channel Convolutional Neural Network (TC-CNN); Second, Two Channel Autoencoder (TC-AE). Furthermore, human-engineered features are extracted by using two methods: First, Gradient Structural Concavity (GSC); Second, Scale Invariant Feature Transform (SIFT). Experiments are performed by complementing one of the HEF methods with one ALF method on 150000 pairs of samples of the word ``AND" cropped from handwritten notes written by 1500 writers. Our results indicate that HDL architecture with AE-GSC achieves 99.7\% accuracy on seen writer dataset and 92.16\% accuracy on shuffled writer dataset which out performs CEDAR-FOX, as for unseen writer dataset, AE-SIFT performs comparable to this sophisticated handwriting comparison tool.\\

\end{abstract}


\section{Introduction}
Handwriting comparison is a task to find the likelihood of similarity between the handwritten samples of the known and questioned writer. This comparison is based on the hypothesis that every individual has peculiar handwriting. Furthermore, the exclusive nature of individual’s handwriting helps differentiate between handwritten samples of distinct individuals. State-of-the-art handwriting analysis tool, CEDAR-FOX was developed at Center of Excellence for Document Analysis and Recognition, University at Buffalo to validate the idea of writer individuality. Srihari et al. in Individuality of Handwriting \cite{Individuality:1} compares the intra-writer feature variation with inter-writer feature variations for the task of handwriting verification and identification. The features used to find these variations were conventional and computational features. Conventional features were extracted using twenty-one rules of discriminating elements of handwriting \cite{intro1:12} \cite{intro2:13}\cite{intro3:14}. Computational features were computed by algorithms and comprised of macro and micro features. Eleven macro features were used to describe paragraph, line, word and character level features which closely resembles five conventional features: pen pressure, writing movements, stroke formation, slantness and height. Micro features were used to describe character and allograph level features. Gradient Structural Concavity feature (GSC) algorithm was used to compute 512 micro binary feature vector. CEDAR-FOX uses the feature variations to compute log likelihood ratio (LLR) for a given pair of input handwritten samples. The results achieved by CEDAR-FOX were state-of-the-art for verification task. Since then there has been significant efforts to improve the accuracy of the writer verification model by bridging the gap between human understanding  and feature representation of an image.

The main contribution of our work is a new hybrid feature set obtained by unifying the handcrafted features from SIFT and deeply learned features from twin Auto-Encoder. 
\begin{figure*}[htbp]
\center\includegraphics[width=18cm,height=2.91112890312cm]{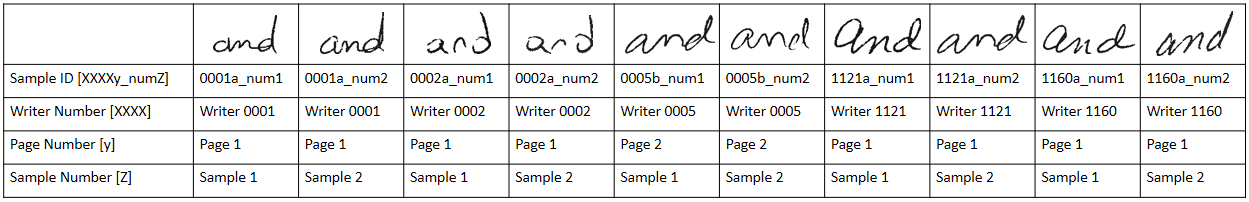}
\caption{Dataset Description}
\label{fig}
\end{figure*}
In the first hybrid feature set, we have used SIFT \cite{Sift:4} as a handcrafted feature extractor. SIFT has found tremendous application in the area of computer vision, image retrieval and recognition task, but has been underutilized in the area of document analysis as suggested by \cite{Siftmore:16}. SIFT captures point level features and comprises of key-point descriptors. Note that it is not our aim to apply entire SIFT based method which mostly relies on BoW methods for generating fixed size feature vector from variable size keypoint descriptors. Instead we use nearest neighbour matching algorithm to map and compare similar key-points between the given two handwriting samples. The resulting mapped features forms one half of our feature set.

The other half is contributed by deeply learned features extracted by Convolutional Neural Networks (CNN). CNNs are often compared to human brain cortex because of hierarchical architecture and richer feature set. CNN has provided state-of-the-art performance in several vision task (e.g document recognition, image classification, object detection, etc). Deep learning networks using CNN have found application in various verification task such as signature verification (e.g SigNet \cite{SigNet:19}) and face verification \cite{FaceVerify:20}. For the task of handwriting comparison we have implemented a baseline architecture for feature extraction using Siamese Network \cite{Siamese:2} which is essentially a Two Channel-CNN (TC-CNN) network. Furthermore, we have implemented an advanced deep learning model, Two-Channel Auto-Encoder (TC-AE) \cite{AE:21} in which the latent representation forms the basis of our feature set.

    The deeply learned features extracted by using TC-CNN/TC-AE are appended to the handcrafted features extracted using SIFT in the first hybrid setting and to the rule based GSC features in the second hybrid setting. Complementary CNN and SIFT \cite{HDLcomplement:7} shows that in general, CNN complements SIFT. Furthermore, \cite{vijay:22} shows that although CNN achieves decent performance, we cannot always infer that CNN will outperform SIFT. Hence, the hybrid deep learning (HDL) model was inspired to capture richer features by combining deep learning with handcrafted features as done by \cite{HDLsecond:8} \cite{CNNmeet:9}.  Experiments are performed to compare the accuracy of SIFT, GSC, TC-CNN, TC-AE and HDL on CEDAR letter dataset. Furthermore, we have validated our results using batch verification functionality of CEDAR-FOX.

\section{Dataset}
Our dataset comprises of ``AND" images extracted from CEDAR Letter dataset \cite{Individuality:1}. The CEDAR dataset consists of handwritten letter manuscripts written by 1567 writers. Each of the writer has copied a source document thrice. Hence, there are a total of 4701 handwritten letter manuscripts. Each letter has vocabulary size of 156 words (32 duplicate words, 124 unique words) and has one to five occurrences of the word ``AND" (cursive and hand printed). Image snippets of the word ``AND" were extracted from each of the manuscript using transcript-mapping function of CEDAR-FOX \cite{TranscriptMapping:3}. The total number of ``AND" image fragments available after extraction are 15,518. Figure 1. shows examples of the ``AND" image fragments.

\subsection{Choice of Dataset}
Our motivation for using the word ``AND" for the handwriting comparison task is based on two premises:
\begin{itemize}
\item Abundance in English Language: The conjunction ``AND" ranks 4th in the list of most frequently used words by Corpus of Contemporary American English (COCA).
\item Availability in CEDAR dataset: On an average there are 10 samples of the word ``AND" from each of the writer. These samples are good enough for comparing the intra-class variations within the same writer and the inter-class variation between different writers.
\end{itemize}

\subsection{Data Inconsistency}
Improper transcript-mapping of ``AND" fragments from manuscript would lead to  inconsistencies in the dataset. For example, handwritten symbols extracted by the transcript-mapping tool may produce outliers as ``AND" words which would lead to reduced accuracy of the overall system. To avoid data inconsistency, outlier symbols are removed at the data validation step by rejecting incorrect input symbols.

\subsection{Data Preprocessing}
 Every handwritten image is padded uniformly on all the four edges so as to have consistent size corresponding to the maximum width and height (384x384) across all the samples. We then downscale the image by a factor of six resulting in a square image of size 64x64.

\subsection{Data Partitioning}
We have compared three approaches for partitioning the dataset for training and testing. Each of these three approaches described below would have an impact on the training and testing accuracy as shown in the result section:
\begin{itemize}
\item Unseen Writer Partitioning:  In this method there exists no writer which is present in both the training (Tr) and testing (Ts) writer set simultaneously. Hence, any test writer would not be a part of training set and vice-versa.
\begin{equation}
T_{r} \bigcap T_s = \emptyset \label{eq}
\end{equation}
\item Shuffled Writer Partitioning: In this method, entire dataset is first shuffled. Hence, there are X writers which are concurrent in both the training (Tr) and testing (Ts) writer  set. Hence, given a test writer may or may not be present in the training set.
\begin{equation}
T_{r} \bigcap T_s = X \label{eq}
\end{equation}
\item Seen Writer Partitioning: In this method, we train over 80\% of each writer’s samples and test over the remaining 20\% samples of each writer.
\begin{equation}
T_{s} =  \bigcup_{j = 1}^{N} 0.2 * S_{j}\label{eq}
\end{equation}
\begin{equation}
T_{r} \bigcup T_s = S \label{eq}
\end{equation}
\end{itemize}

\section{Deep Learning}
Our approach is based on using CNN as a feature extractor. Recently, many popular CNN architectures have been employed for feature extraction (e.g VGG \cite{VGG:23}, Alexnet \cite{Alexnet:24}, Resnet \cite{Resnet:25}, Inception \cite{Inception:26}). Since our goal is to focus on testing baseline models, we have designed a simplistic five layer CNN model for high level feature extraction. At the same time, since the AND images are in grayscale with minimal pixel, hence, we remove the  max pooling layer so that we do not lose any important datapoint. However, we use a valid padding in CNN with larger receptor size to reduce the dimensionality. Furthermore we expand our networks to learn the features of two input images simultaneously.

\begin{figure}[htbp]
\center\includegraphics[width=9cm,height=2.16131386861cm]{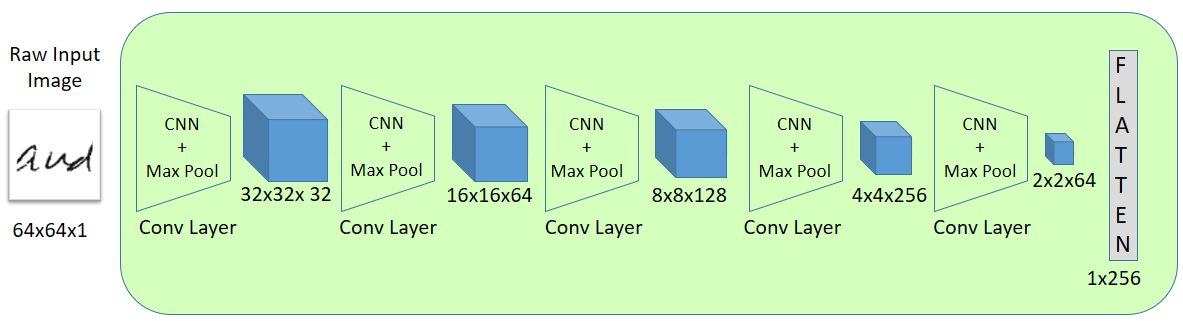}
\caption{CNN Architecture}
\label{fig}
\end{figure}

\subsection{Two channel CNN (TC-CNN)}
Most commonly a two channel CNN network, is called as Siamese Network named after the work done by Bromley,  et al \cite{Siamese:2}, where, they use an energy function to compute their loss. We have a similar setting where the two networks share weights. Essentially the same network is responsible to generate features for the two different images ($I_{i}, I_{j}$)  in a parallel setting. We do not use energy function to calculate our loss, but we experiment on multiple functions to compare the features $(F_{i}, F_{j})$  from $I_{i}$ and $I_{j}$. In the first setup we tie $(F_{i}, F_{j})$ by concatenating them $(F_{x}\_conc = F_{i} \bigcup F_{j})$, and then train $F_{x}\_conc$ over Fully Connected layers. In the second setup we find the difference $(F_{x}\_diff)$ between $(F_{i}, F_{j})$. This feature difference vector is then used as an input and trained over Fully Connected layers. Finally we obtain the output in the form of a two class classifier and hence categorical cross entropy (CCE) loss function after the last softmax layer becomes an obvious choice. We design a simplistic model containing five Convolution layers each followed by a max pooling layer as displayed in Figure 2. Two gray scale images of height h=64 and width w=64 are input to the first convolution layer containing 32 kernels k of size 3x3. Max pool layers of stride 2x2 are introduced to reduce the h to h/2 and w to w/2 after each convolution. We double the size of k until the fourth conv layer. For the fifth conv layer k=64 so that post flatten the total number of neurons output from one channel is 256.

\begin{figure}[htbp]
\center\includegraphics[width=9cm,height=2.33660377358cm]{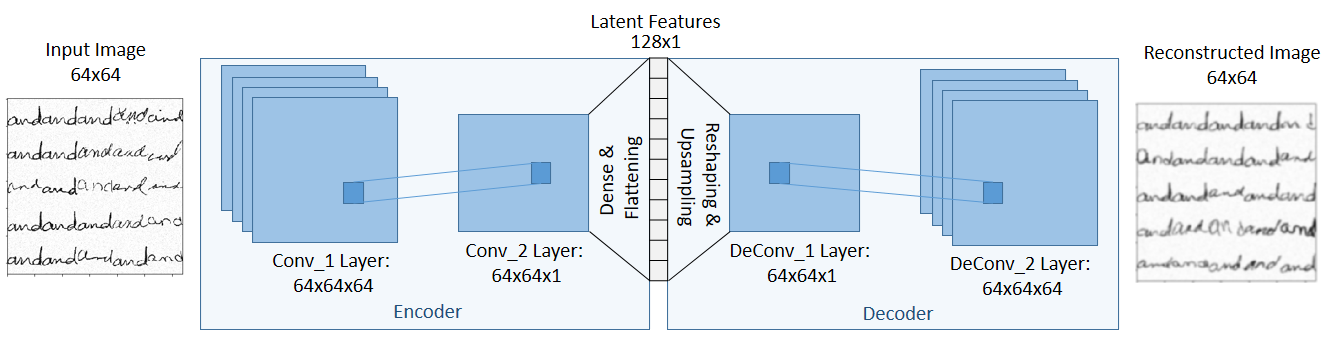}
\caption{Auto Encoder Architecture}
\label{fig}
\end{figure}

\subsection{Two channel Autoencoder (TC-AE)}
Although CNN is an excellent feature extractor, it maps the images to only the class they belong, which in this case would be great if our task was a writer identification task. In CNN the difference between an output class and the actual class is propagated back and the weights are updated to only have a better mapping between $F_{i},F_{j}$ and class $C_{ij}$ (Where $C_{ij}$ belongs to {0,1}). However, since the ``and" images are in grayscale, we need more robust features that describe the image, not necessarily mapping any class. We do this by reconstructing the image using an AutoEncoder, and by adding the reconstruction Mean Absolute Error (MAE) between the logistic outputs and the image pixels as a regularizer. Furthermore, we do the same setup, as displayed in TC-CNN, considering the latent variables $L_{i}, L_{j}$ of $I_{i}, I_{j}$ as $F_{i}, F_{j}$. We train the AE for a batch b = 512 of images that appear in each channel one by one. The encoder architecture contains two Convolution layers with k=64 and k=1 respectively, followed by three fully connected (FC) layers with neurons n=4096, n=1024, n=128 respectively, as shown in Figure 3. In the decoder architecture the encoder is simply reversed to maintain balance between encoder and decoder, however, we apply upsampling with cubic interpolation (CI) after the Convolution layers to resize the 2D arrays to size 64x64. The FC layer in encoder with n = 128 acts as latent representation of the input. In one training iteration the weights AE are updated, then encoder is duplicated to generate two channels. Each channel outputs feature vectors $F_{i}$ and $F_{j}$ of length 128. Finally, the same operations are performed over $F_{i}$ and $F_{j}$ as performed in the two setups of TC-CNN.

\begin{figure}[htbp]
\includegraphics[width=8.5cm,height=5.39277010561cm]{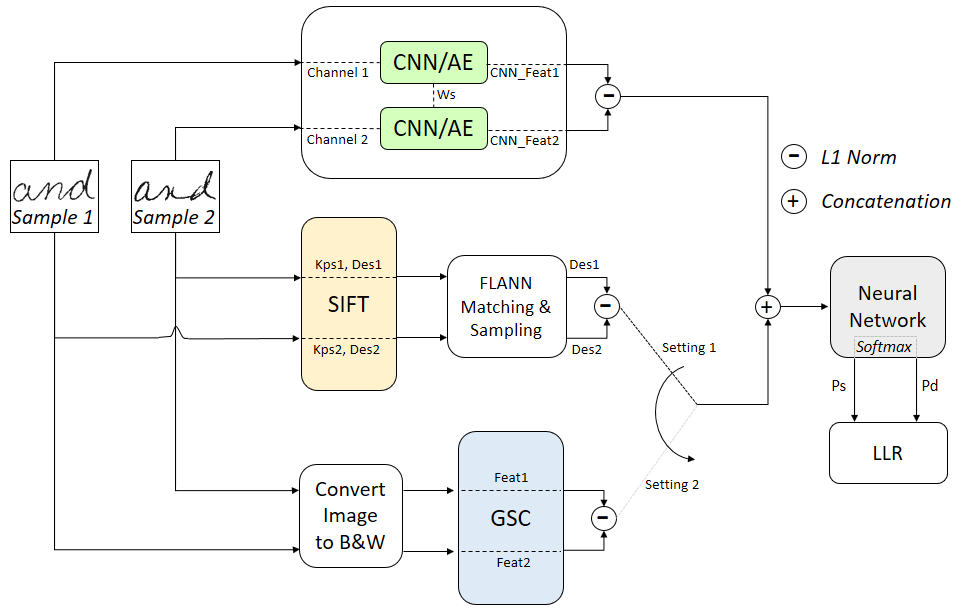}
\caption{Hybrid Deep Learning Architecture}
\label{fig}
\end{figure}

\section{Hybrid Deep Learning}
In this section, we describe in detail the two hybrid deep learning techniques to solve for handwriting comparison task:

\begin{figure}[htbp]
\center\includegraphics[width=8cm,height=4.01384083045cm]{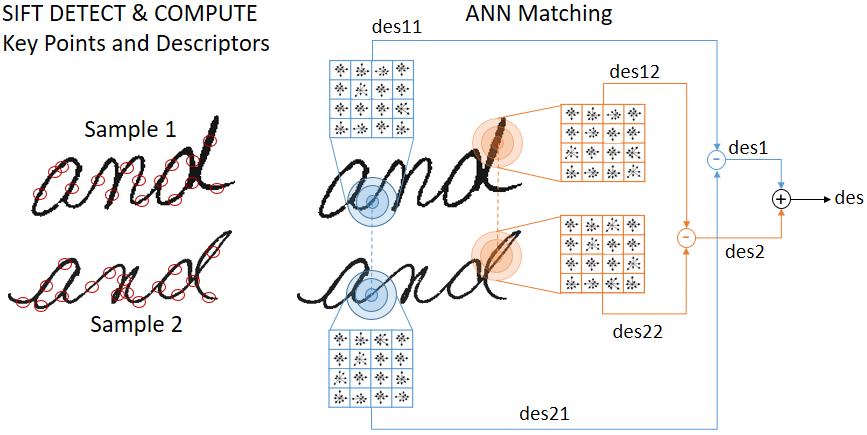}
\caption{SIFT Feature Extractor with FLANN Feature Matching }
\label{fig}
\end{figure}

\subsection{TC-AE/TC-CNN with SIFT}
In the first hybrid setting we append the handcrafted features from SIFT to deeply learned features from TC-CNN/TC-AE as shown in Figure 4. SIFT outputs variable number of keypoints with descriptors for a given input image sample. Each keypoint is a result of a stable maxima produced across all the differences of blurred scales in an octave. The blur is produced by convolving the image with a gaussian kernel. The outliers from the resulting keypoints are removed using low contrast detector and Harris corner detector. Associated with each input sample are variable number of keypoints. For each keypoint SIFT creates scale, rotational and orientational invariant image descriptors. Length of each descriptor is 128. We use nearest neighbour mapping algorithm FLANN \cite{FLANN:10} to match n keypoints for given pair of input image samples as shown in Figure 5. The matching process is similar to forensic document examination wherein the handwriting samples are matched on the basis of: strokes (connecting, beginning, ending), slantness, flourishments and embellishments. After the matching process, we take the L1 difference between the n matched descriptors. The resulting feature vector is of fixed size n*128 which represents the differences between the features of the two handwriting samples. This difference feature vector of SIFT is appended to the deeply learned features of TC-CNN/TC-AE. Combined features are then fed as an input to a two class neural network classifier. We use the softmax layer of the neural network classifier to provide a degree of similarity (LLR) between the pair of input samples. LLR is computed by taking the logarithm of the ratio of the probability of  input pairs being similar to the probability of the pair being dissimilar. $x_{1}$ and $x_{2}$ represent the features of image sample 1 and 2 respectively while $c_{0}$ and $c_{1}$ represent similar and dissimlar classes. Hence, the LLR equation is computed as follows: 

\begin{equation}
\begin{split}\label{eq}
LLR & = \frac{P(x_{1},x_{2}|c_{0})}{P(x_{1},x_{2}|c_{1})}\\
      & = \frac{\frac{P(c_{0}|x_{1},x_{2})*P(x_{1},x_{2})}{P(c_{0})}}{\frac{P(c_{1}|x_{1},x_{2})*P(x_{1},x_{2})}{P(c_{1})}}\\
     & = \frac{P(c_{0}|x_{1},x_{2})*P(c_{1})}{P(c_{1}|x_{1},x_{2})*P(c_{0})}\\
\end{split}
\end{equation}

Under the fair assumption that the probability of input samples being similar and dissimilar is the same i.e $P(c_{0}) = P(c_{1})$. We can safely conclude that:
\begin{equation}
LLR = \frac{P(c_{0}|x_{1},x_{2})}{P(c_{1}|x_{1},x_{2})}\label{eq}
\end{equation}

Hence, for computing LLR we use logarithmic ratio of the resultant softmax probabilities for each class within the neural network classifier. 
\begin{table*}[]
\centering
\caption{Experimental results of comparison between Deep Learning, Handcrafted and HDL methods. Here notation "f1 - f2" means the results belong to setup when network was trained on the vector difference of the features obtained, from the two input samples, using the methods in the third row of column header. And $``f1 \bigcup f2"$ means the results belong to setup when network was trained on the union of the features obtained, from the two input samples, using the methods in the third row of column header. The first rowc code of column header tells the category of feature extraction.}
\label{my-label}
\begin{tabular}{|c|c|c|c|c|c|c|c|c|c|}
\hline
\rowcolor[HTML]{EFEFEF} 
\cellcolor[HTML]{F7F7F7}\textbf{Method} & \multicolumn{2}{c|}{\cellcolor[HTML]{EFEFEF}\textbf{Deep Learning}} & \multicolumn{2}{c|}{\cellcolor[HTML]{F7F7F7}\textbf{Handcrafted}}            & \multicolumn{4}{c|}{\cellcolor[HTML]{EFEFEF}\textbf{Hybrid Deep Learning}}    & \cellcolor[HTML]{F7F7F7}\textbf{Baseline}                \\ \hline
\multicolumn{10}{|c|}{\textit{\textbf{Seen dataset Partitioning}}}                                                                                                                                                                                                                                                                      \\ \hline
\rowcolor[HTML]{EFEFEF} 
\cellcolor[HTML]{F7F7F7}\textbf{Setups} & \textbf{CNN}                      & \textbf{AE}                     & \cellcolor[HTML]{F7F7F7}\textbf{SIFT} & \cellcolor[HTML]{F7F7F7}\textbf{GSC} & \textbf{CNN\_SIFT} & \textbf{AE\_SIFT} & \textbf{CNN\_GSC} & \textbf{AE\_GSC} & \cellcolor[HTML]{F7F7F7}\textbf{CEDAR-FOX}               \\ \hline
\rowcolor[HTML]{EFEFEF} 
\cellcolor[HTML]{F7F7F7}f1 - f2           & 93.47                             & 93.21                           & \cellcolor[HTML]{F7F7F7}71.49         & \cellcolor[HTML]{F7F7F7}87.94        & 84.9               & 86.47             & 95.35             & \textbf{96.11}   & \cellcolor[HTML]{F7F7F7}                                 \\ \cline{1-9}
\rowcolor[HTML]{EFEFEF} 
\cellcolor[HTML]{F7F7F7}f1 $\bigcup$ f2           & 98.78                             & 99.56                           & \cellcolor[HTML]{F7F7F7}76            & \cellcolor[HTML]{F7F7F7}97.38        & 78.5               & 82.39             & 98.95             & \textbf{99.78}   & \multirow{-2}{*}{\cellcolor[HTML]{F7F7F7}81.36}          \\ \hline
\multicolumn{10}{|c|}{\textit{\textbf{Shuffled dataset Partitioning}}}                                                                                                                                                                                                                                                                  \\ \hline
\rowcolor[HTML]{EFEFEF} 
\cellcolor[HTML]{F7F7F7}f1 - f2           & 84.21                             & 87.14                           & \cellcolor[HTML]{F7F7F7}70.15         & \cellcolor[HTML]{F7F7F7}89.65        & 72.41              & 76.17             & 89.05             & \textbf{91.43}   & \cellcolor[HTML]{F7F7F7}                                 \\ \cline{1-9}
\rowcolor[HTML]{EFEFEF} 
\cellcolor[HTML]{F7F7F7}f1 $\bigcup$ f2           & 86.8                              & 91.23                           & \cellcolor[HTML]{F7F7F7}71.1          & \cellcolor[HTML]{F7F7F7}89.91        & 70.21              & 70.21             & 90.29             & \textbf{92.16}   & \multirow{-2}{*}{\cellcolor[HTML]{F7F7F7}82.53}          \\ \hline
\multicolumn{10}{|c|}{\textit{\textbf{Unseen dataset Partitioning}}}                                                                                                                                                                                                                                                                    \\ \hline
\rowcolor[HTML]{EFEFEF} 
\cellcolor[HTML]{F7F7F7}f1 - f2           & 53.48                             & 55.96                           & \cellcolor[HTML]{F7F7F7}59.21         & \cellcolor[HTML]{F7F7F7}52.69        & 61.9               & \textbf{64.24}    & 58.32             & 58.31            & \cellcolor[HTML]{F7F7F7}                                 \\ \cline{1-9}
\rowcolor[HTML]{EFEFEF} 
\cellcolor[HTML]{F7F7F7}f1 $\bigcup$ f2           & 51.24                             & 56.9                            & \cellcolor[HTML]{F7F7F7}58.55         & \cellcolor[HTML]{F7F7F7}51.89        & 62.21              & \textbf{63.12}    & 55.89             & 59.91            & \multirow{-2}{*}{\cellcolor[HTML]{F7F7F7}\textbf{80.92}} \\ \hline
\end{tabular}
\end{table*}

\begin{figure*}[htbp]
\center\includegraphics[width=15cm,height=4.87056213018cm]{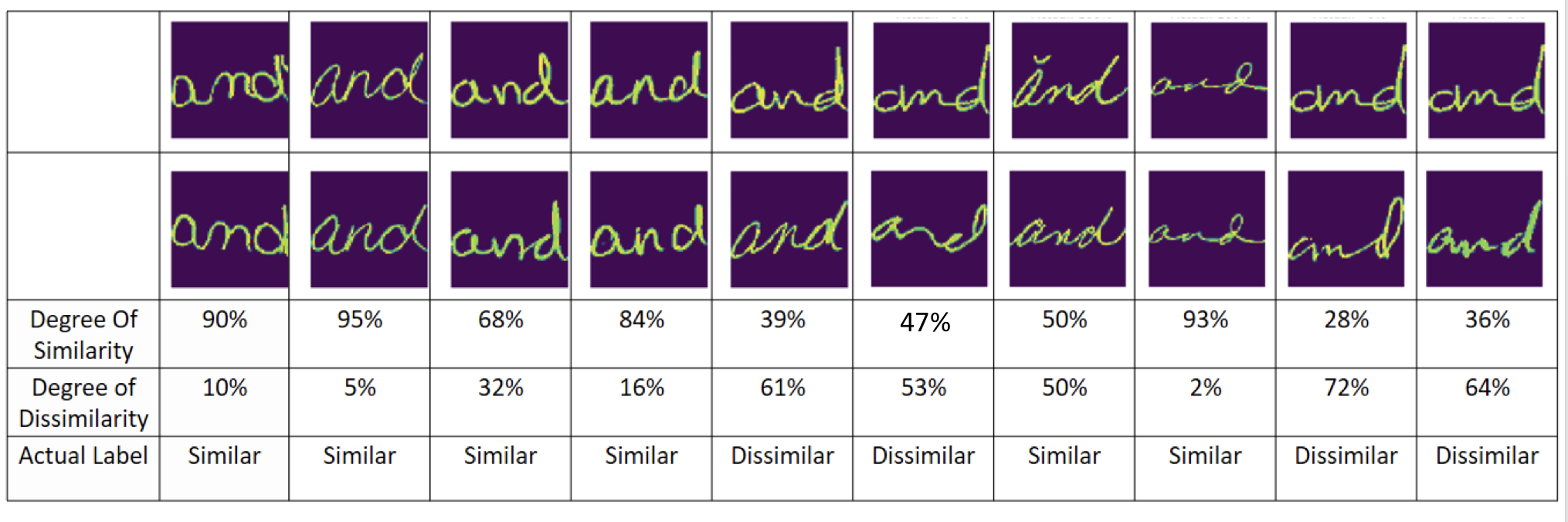}
\caption{Sample Input and Output using HDL}
\label{fig}
\end{figure*}

\subsection{TC-AE/TC-CNN with GSC}
In the second hybrid setting we append the rule based features obtained from GSC to deeply learned features obtained from  TC-CNN/TC-AE. Since GSC algorithm requires binarized images as input, we use thresholding technique to binarize the images. First step of GSC computation is to uniformly subsample the input image into 4x4 parts. Each part comprises of three features:
\begin{itemize}
\item Gradient Features are computed by first convolving 3x3 sobel filter \cite{Sobel:11} with each subsample to obtain gradient angle at each pixel with respect to all its adjacent pixels. The gradient angle would vary between 0 to 2$\pi$ radians in polar coordinate system. The polar coordinate system is now discretized into 12 bins, each of size $2 \pi /12$ radians. We now find the frequency of gradient angles occuring in each bin within each subsample. If the frequency of a bin is greater than the set threshold then the value of the binary gradient feature vector for this bin is 1. If the frequency is lower than the threshold than the value for the bin is set to 0. Hence, each subsample would contribute to a binary gradient feature vector of size 12. The entire image would have 12*4x4=192 binary gradient features.
\item Structural Features are computed by applying set of 12 rules to each pixel with respect to its adjacent eight neighbours within each subsample. We use standard 12 rules as used by John T. Fatava and Geetha Srikantan \cite{GSCfirst:5} for finding structural features which detects horizontal line, vertical line, diagonal rising, diagonal falling and four corners. Each rule is considered a bin. Hence, structural feature vector has 12 bins. Similar to gradient feature bins, thresholding structural feature bins would result in a binary vector. The entire image would have 12*4x4=192 binary structural features.
\item Concavity Feature are of size 128 bits composed of three subfeatures:
\begin{itemize}
\item Pixel Density subfeatures are the number of black pixels within each subsample. Thresholding the pixel density would result into a binary value. Entire image would have 1*4x4=16 binary pixel density subfeatures.
\item Large-Stroke features are computed by finding the largest continuous sequence of black pixels along horizontal and vertical directions. Thresholding the sequence would result in a binary value. Entire image would have 2*4x4 large stroke subfeatures.
\item Concavity features capture up/down/left/right pointing concavities by convolving with starlike operator \cite{GSCfirst:5}.  The algorithm to compute the star operator is beyond the scope of this paper. Rules are associated to detect the concavities. Entire image would have 5*4x4 = 80 concavity subfeatures.
\end{itemize}
\end{itemize}

We use L1 difference between the pairs of GSC features of handwritten samples. The difference feature vector is then appended to the deeply learnt features of TC-CNN/TC-AE. Similar to SIFT the combined features are then fed as an input to a two class neural network classifier. The classifier outputs a degree of similarity between pair of input handwritten samples as described in Equation 5.

\section{Experiment and Results}
Our experimental setup comprises of 3 parts: Setting up SIFT extractor; setting up GSC extractor; setting up HDL models. For feature extraction using SIFT/GSC; we pre-extract the matching features for each pair of images in Tr\_set and Ts\_set of each dataset and store it on file system (FS) in CSV formats with corresponding name. For HDL we train and test our model using four 11GB NVIDIA GTX 1080 Ti GPUs and a TensorFlow backend. We consider CEDAR-FOX (CF) software as our baseline for evaluating our architectures. We input pairwise combination of image files to CF’s batch verification tool. CF generates an excel with negative and positive scores of LLR, where the higher the positive value means a higher similarity and lower the negative value indicates a higher dissimilarity. This process is followed for ts\_set of each dataset results are compared to the results obtained from our models. We train our model for over 4000 epochs. The results are presented in table1.

The code for GSC feature extraction, SIFT feature extraction using OpenCV and the code for HDL training are available on github at the following link: \url{https://github.com/mshaikh2/HDL\_Forensics.git}

The accuracy of a model is defined as the ability of system to generate a positive LLR if the ground truth images (GT) are labeled similar i.e 1 and negative if the GT images are labelled dissimilar i.e 0. The results in table 1 represent the percentage accuracy obtained using each architecture with the mathematical setups, $f_{1}$ + $f_{2}$ and $f_{1}$ - $f_{2}$. Furthermore there are many measures for finding similarity between features like euclidean, hamming, chi square distance, however, our focus in this paper is to compare the similarity based features with unified features. Hence, we choose to compare simple difference based features with concatenated based features. The results on unseen writer dataset are relatively worse as the writing in the testing set was totally different from what the model saw during training, however results on seen and shuffled dataset seems to outperform and are comparable to the baseline CEDAR-FOX results. Specifically the HDL architecture using AE and GSC under the concatenated setting performs the best in shuffled dataset, where as, the HDL model using AE and SIFT performed best in unseen writer dataset. Overall, the performance of concatenation of features performs better than similarity based features as we can argue the loss of information in the latter, which is clearly evident by our results.

\section{Conclusion}
Overall for handwriting comparison task, HDL proves to be a promising architecture and provides decent accuracy even on unseen writer dataset, based on evaluation of results against CEDAR-FOX. However, extracting features from SIFT and GSC is a time consuming task and it is worth noting that there is a scope to learn these features automatically. It is also evident that human engineered feature extractors like SIFT and GSC complement the deeply learned features extracted by a CNN. However, feature extraction using SIFT and GSC add about 70\% to the training time, and create features from different sources, which makes it inconvenient to train the model end to end. We open a new realm of research in the field of Handwriting Comparison for bridging the gap between these feature extraction methods as a future scope, which can lead to increase in overall accuracy of the system.

\section{Acknowledgement}
We are extremely grateful to CEDAR for providing ac-
cess to CEDAR-FOX for validating the results obtained by hybrid deep learning models. We would also like to thank Dr. Mingchen Gao for providing GPU compute environment. Finally, we thank our team mates for proof reading our paper.

\bibliography{ref1} 
\bibliographystyle{ieeetr}

\end{document}